\documentclass[10pt,twocolumn,letterpaper]{article}

\usepackage{cvpr}  
\usepackage{amsmath,amssymb,amsfonts}
\usepackage{algorithmic}
\usepackage{graphicx}
\usepackage{textcomp}
\usepackage{breqn}
\usepackage{xcolor}
\usepackage{float}
\usepackage{url}
\usepackage{caption}
\usepackage{multirow}
\usepackage{subcaption}
\usepackage{booktabs}
\usepackage{placeins}
\def\BibTeX{{\rm B\kern-.05em{\sc i\kern-.025em b}\kern-.08em
    T\kern-.1667em\lower.7ex\hbox{E}\kern-.125emX}}

\definecolor{cvprblue}{rgb}{0.21,0.49,0.74}
\usepackage[pagebackref,breaklinks,colorlinks,citecolor=cvprblue]{hyperref}

\def\confName{CVPR}
\def\confYear{2024}

\title{FPN-IAIA-BL: A Multi-Scale Interpretable Deep Learning Model for Classification of Mass Margins in Digital Mammography
}

\author{Julia Yang\\
Duke University \\
Durham, NC, USA \\
{\tt\small julia.yang@duke.edu}
\and
Alina Jade Barnett\\
Duke University\\
Durham, NC, USA \\
{\tt\small alina.barnett@duke.edu}
\and
Jon Donnelly\\
Duke University\\
Durham, NC, USA \\
{\tt\small jon.donnelly@duke.edu}
\and
Satvik Kishore\\
Duke University\\
Durham, NC, USA \\
{\tt\small satvik.kishore@duke.edu }
\and
Jerry Fang\\
Duke University\\
Durham, NC, USA \\
{\tt\small jerry.d.fang@alumni.duke.edu}
\and
Fides Regina Schwartz\\
Brigham and Women’s Hospital\\
Boston, MA, USA \\
{\tt\small frschwartz@bwh.harvard.edu}
\and
Chaofan Chen\\
University of Maine\\
Orono, ME, USA \\
{\tt\small chaofan.chen@maine.edu}
\and
Joseph Y. Lo\\
Duke University\\
Durham, NC, USA \\
{\tt\small joseph.lo@duke.edu}
\and
Cynthia Rudin\\
Duke University\\
Durham, NC, USA \\
{\tt\small cynthia@cs.duke.edu}
}

\begin{document}
\maketitle

\begin{abstract}
Digital mammography is essential to breast cancer detection, and deep learning offers promising tools for faster and more accurate mammogram analysis. 
In radiology and other high-stakes environments, uninterpretable (``black box'') deep learning models are unsuitable and there is a call in these fields to make interpretable models. 
Recent work in interpretable computer vision provides transparency to these formerly black boxes by utilizing prototypes for case-based explanations, achieving high accuracy in applications including mammography. However, these models struggle with precise feature localization, reasoning on large portions of an image when only a small part is relevant. This paper addresses this gap by proposing a novel multi-scale interpretable deep learning model for mammographic mass margin classification. Our contribution not only offers an interpretable model with reasoning aligned with radiologist practices, but also provides a general architecture for computer vision with user-configurable prototypes from coarse- to fine-grained prototypes.
\end{abstract}

\section{Introduction}

\begin{figure}
    \centering
    \includegraphics[width=0.9\linewidth]{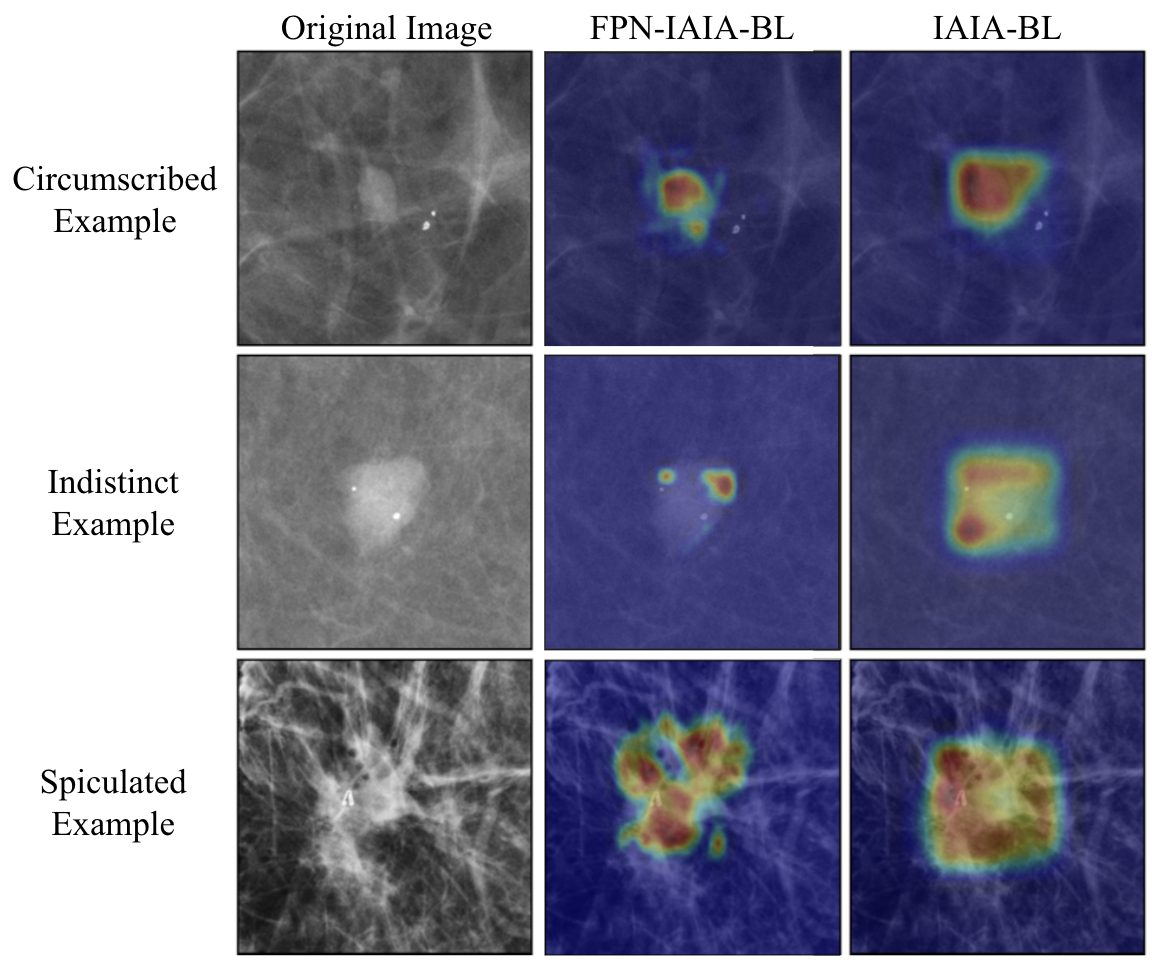}
    \caption{\textbf{Activation maps for FPN-IAIA-BL in comparison to IAIA-BL.} FPN-IAIA-BL can learn human interpretable prototypes at any scale, including fine-grained details most salient to mass margin classification.}
    \label{fig:3by3}
\end{figure}

Digital mammography plays a crucial role in detecting and diagnosing breast cancer, a pervasive health concern worldwide. Advancements in deep learning and computer vision have increased the speed and accuracy of lesion classifications for mammography. However, when used for high-stakes tasks like medical diagnoses, deep learning models should be inherently interpretable so that, among other advantages, models can be ``fact checked'' \cite{rudin2019stop}. 

Recent work has shown that interpretable, case-based machine learning models can provide accurate, human understandable explanations for their predictions while performing on par with other state-of-the-art models \cite{LiLiuChenRudin, PPNet}. 
These prototype-based deep learning models have also been applied to digital mammography by Barnett et al. \cite{barnett2021case}, who developed the Interpretable AI Algorithm for Breast Lesions (IAIA-BL) model, an interpretable model for \textit{mass margin} classification. They focused on classification on \textit{margins}, a descriptor of the edges around the mass, because it is a key factor in identifying cancerous lesions under the Breast Imaging Reporting and Data System (BI-RADS).
IAIA-BL successfully classified margins using prototypes, as shown in the third column of Figure \ref{fig:3by3}. However, the prototypes often identified more than just the margin or even the entire lesion, leaving any detailed analysis of the margin to the user.

To address this gap, we develop FPN-IAIA-BL, a multi-scale interpretable deep learning model for mammographic mass margin classification. 
It can be configured to provide prototypes at various levels of granularity, with multiple scales within the same model. We build the model’s architecture using both the Feature Pyramid Network (FPN) and IAIA-BL model. We developed a new training schedule and objective function, as the training methods and loss terms used by these predecessors were insufficient to train the combined architecture. The main contributions of this work are that: 

\begin{itemize}
    \item We develop an inherently interpretable deep learning architecture that learns prototypes at multiple scales.
    \item We train FPN-IAIA-BL, which provides specific prototype activations for mass margin classification. 
\end{itemize}

\section{Related Work} \label{sec:related}
Interpretability of deep learning models is critical for high-stakes applications like breast cancer detection and diagnosis. In recent years, \textit{inherently interpretable} deep neural networks have grown in popularity. As compared to \textit{posthoc explanation} techniques such as saliency visualizations \cite{bach2015pixel,simonyan2013deep,smilkov2017smoothgrad,springenberg2014striving,pmlr-v70-sundararajan17a, ZeilerFe14}, activation maximization \cite{Erhan, Nguyen16, wang21, yoshimura21, Yosinski2015}, and image perturbation methods \cite{fong2019understanding, ivanovs21} which approximate model reasoning after training, \textit{inherently interpretable} techniques such as \cite{ma2024looks, rymarczyk2022protopool, wang2021interpretable, PPNet, donnelly2022deformable, barnett2022interpretable, nauta2021neural, LiLiuChenRudin, barnett2022mapping} provide explanations guaranteed to be faithful to the model's underlying decision-making process. 

FPN-IAIA-BL uses inherently interpretable case-based reasoning with prototypes by building upon IAIA-BL \cite{barnett2021case}, a case-based model for mass margin classification. IAIA-BL was limited to learning prototypes at only one scale, with prototypes often identifying more of the image than is relevant for margin classification. 
In contrast, FPN-IAIA-BL learns prototypes at various scales including highly-localized, fine-grained prototypes that select small details, as shown in Figure \ref{fig:3by3}. This is possible because FPN-IAIA-BL incorporates features at various scales.

Typically, a key challenge in mammogram analysis is capturing information at various scales, since traditional CNN architectures focus on a single image resolution. Multi-scale approaches like \cite{cui2016multiscale} and \cite{lin2017fpn} address this challenge by incorporating features extracted at different scales within the network. A foundational architecture for multi-scale predictions is the Feature Pyramid Network (FPN) \cite{lin2017fpn} which introduces a bottom-up and top-down pyramidal architecture that produces multiple feature maps from fine-grained to coarse. As a result, FPN's are able to localize to objects of multiple scales for object detection.

Our FPN-IAIA-BL architecture leverages this bottom-up and top-down pyramidal architecture to learn prototypes at multiple scales by augmenting IAIA-BL's VGG-16 backbone with a similar structure, detailed in Section \ref{sec:architecture}. Furthermore, our model also provides visual, human interpretable, case-based reasoning for each classification.

\section{FPN-IAIA-BL Architecture} \label{sec:architecture}
\begin{figure*}
    \centering
    \includegraphics[width=0.8\textwidth]{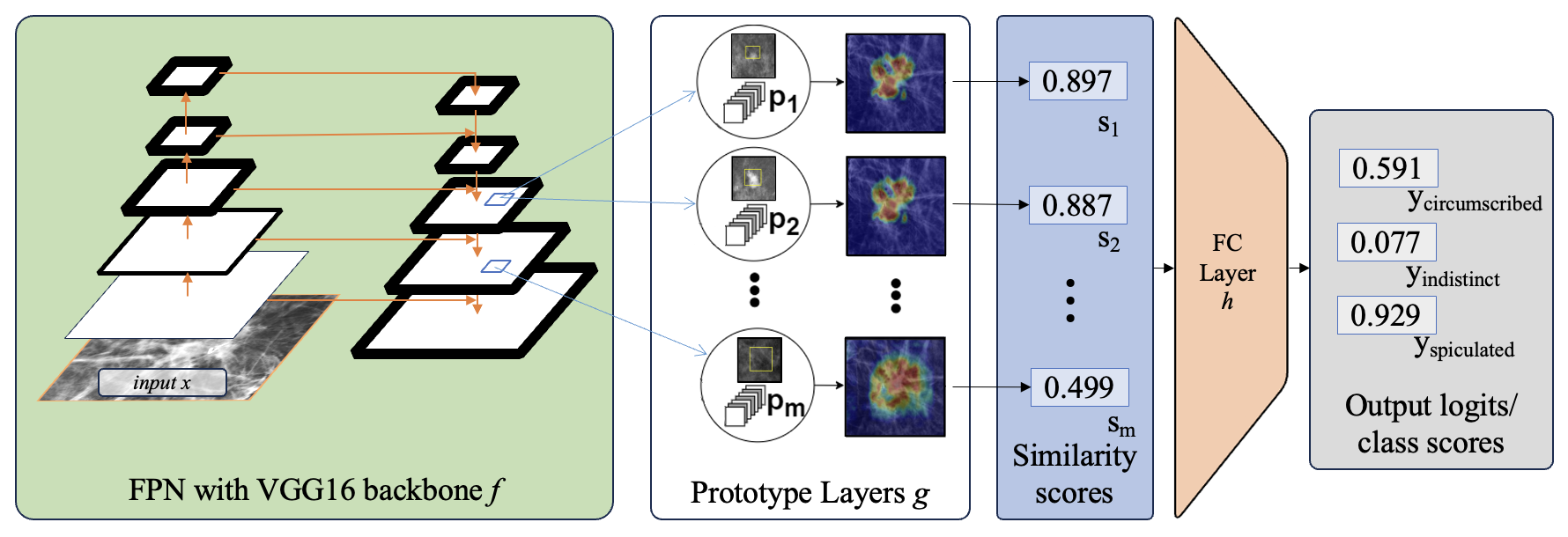}
    \caption{\textbf{FPN-IAIA-BL Architectre.} The input image $\mathbf{x}$ passes through convolutional layers $f$ consisting of an FPN with a VGG-16 backbone, which creates an pyramid of feature maps $f(\mathbf{x})$. Each patch of each level of the feature pyramid (referred to as FPN level) is then compared to each prototype of the same FPN level using a cosine distance to produce an activation map. 
    The activation map is then used to calculate an overall similarity score $s_j$ between the input image and the prototype for each prototype. Finally, a set of fully connected last layer produces logits $y_{\text{margin}}$ for each margin class. }
    \label{fig:architecture}
\end{figure*}

Inspired by the Feature Pyramid Network (FPN), the FPN-IAIA-BL model adds lateral and top-down connections to the original VGG-16 convolutional layers used in the IAIA-BL architecture as its foundation. Figure \ref{fig:architecture} illustrates this architecture. The model consists first of an FPN that extracts useful feature maps at multiple scales, allowing for more varied representation than single-scale IAIA-BL. The FPN is followed by the prototype layer $g$ in which the input image's feature maps are compared to learned prototypes to produce similarity scores. Fully connected layer $h$ then uses the similarity scores to produce margin class predictions. 
\subsection{Multi-Scale Feature Maps from Feature Pyramid Network}
IAIA-BL \cite{barnett2021case} uses a CNN to create a single feature map $z$ which limits the network to prototypes at the scale of that output feature map. In contrast, FPN-IAIA-BL uses the latent feature maps from multiple layers in the CNN, which have different spatial and semantic scales. Thus, the output of the set of convolutional layers $f$ in FPN-IAIA-BL is a set of feature maps of varying spatial scale, which we refer to as the feature pyramid $f(x) =\mathbf{Z}=\{\mathbf{z}^{(2)},\mathbf{z}^{(3)},\mathbf{z}^{(4)},\mathbf{z}^{(5)}\}$. For our implementation, the coarsest feature maps were 14 by 14, and finest were 56 by 56.

For the VGG-16 backbone, we use the output from each block's max-pooling layer to form the intermediate feature map levels in the bottom-up pathway (left column of backbone in Figure \ref{fig:architecture}). We also include the final output of the convolutional layers as a feature map at the top. We denote these bottom-up feature maps as $\mathbf{C}=\{\mathbf{c}^{(2)}, \mathbf{c}^{(3)}, \mathbf{c}^{(4)}, \mathbf{c}^{(5)}\}$ where $\mathbf{c}^{(2)}$ is the base of the bottom-up pyramid, and $\mathbf{c}^{(5)}$ is the top. 

As in FPN \cite{lin2017fpn}, the top-down pathway produces a second feature pyramid. For each level, an upsampled feature map with spatially coarser information is combined with a corresponding laterally connected feature maps from the bottom-up pyramid. Then, each combined feature map is passed through a $3\times 3$ convolution to reduce the aliasing effect of upsampling and output the feature map $\mathbf{z}^{(l)}$. 

\begin{equation}
\begin{gathered}
\mathbf{z}^{(5)} = \text{Conv1x1}(\mathbf{c}^{(5)}) \\
\mathbf{z}^{(l)} = \text{Conv3x3} \Big(\text{Up}(\mathbf{z}^{(l+1)} ) + \text{Conv1x1}(\mathbf{c}^{(l)})\Big); l \in \{2, 3, 4\}
\end{gathered}
\end{equation}


\subsection{Prototype Layer}\label{subsec:proto}
In the prototype layer $g$,  we have $m$ prototypes where each prototype can be configured to represent a specific class $c$ and FPN level $l$.  For $m$ prototypes, let $S = \{(c_j, l_j, j)\}_{j=1}^m$  represent our prototype configuration, and denote our prototypes as $\mathbf{P} = \{\mathbf{p}^{(c, l, j)}\}_S$  where the $j$-th prototype is from class $c$ with FPN level $l$. Each prototype is $1\times 1\times d$  so that each prototype has the same feature dimension $d$ as the convolutional feature pyramid. As in IAIA-BL \cite{barnett2021case}, the prototypes can be interpreted as a characteristic pattern representing a specific class. It can be visually understood by examining a segment of the training image where this pattern was derived. 

Once we have computed each feature map in the convolutional feature pyramid $f(x),$ we compute the similarity between each prototype in prototype layer $g$ and the corresponding feature map.
The FPN-IAIA-BL similarity score $s_j$ differs from that of IAIA-BL in three ways. 

First, because the prototypes are assigned to specific FPN levels $l$, similarities for a set of prototypes $\mathbf{p}^{(\cdot, l, \cdot)}$ are computed only using the feature map from the same FPN level $\mathbf{z}^{(l)}$. 
Second, instead of using inverted $L_2$ distance based similarity, we use a cosine similarity as described in  \cite{donnelly2022deformable, wang2021interpretable}.  The cosine similarity is calculated between a prototype and each $1\times1\times d$ patch within the corresponding feature map. We denote the patch in a feature map of size $\eta_l \times \eta_l \times d$ as $n \in \{(1,1), \dots, (1, \eta_l), (2, 1), \dots, (\eta_l, \eta_l)\}$. Thus, the cosine similarity for a single patch is:
\begin{equation}
s^{(l)}_{j, n} = \frac{\textbf{z}_n^{(l)}}{||\textbf{z}_n^{(l)}||} \cdot \frac{\textbf{p}^{(c,l,j)}}{||\textbf{p}^{(c,l,j)}||} 
\end{equation}
Third, in order to focus activation on the most salient features in each image, we use focal similarity as introduced in ProtoPool \cite{rymarczyk2022protopool}. Retaining the top-k average pooling from Kalchbrenner et al. \cite{kalchbrenner2014convolutional} and IAIA-BL \cite{barnett2021case}, focal cosine similarity is computed as:
\begin{equation}
\begin{aligned}
g(l, j) = \frac{1}{k}\sum\text{top}_k&(\{s^{(l)}_{j,n}\}_{n=(1,1)}^{\space\space (\eta_l, \eta_l)}) \; - \\
&\frac{1}{\eta_l^2}\sum_{n=(1,1)}^{(\eta_l,\eta_l)}(\{s^{(l)}_{j,n}\}_{n=(1,1)}^{\space\space (\eta_l, \eta_l)})
\end{aligned}
\end{equation}

The last stage of FPN-IAIA-BL is a fully connected layer $h$ which weights the similaritie scores and applies a softmax to predict probabilities for each mass-margin class. 

\section{Data and Training} \label{sec:data}

\begin{figure*} [h!]
    \centering
    \includegraphics[width=0.85\textwidth]{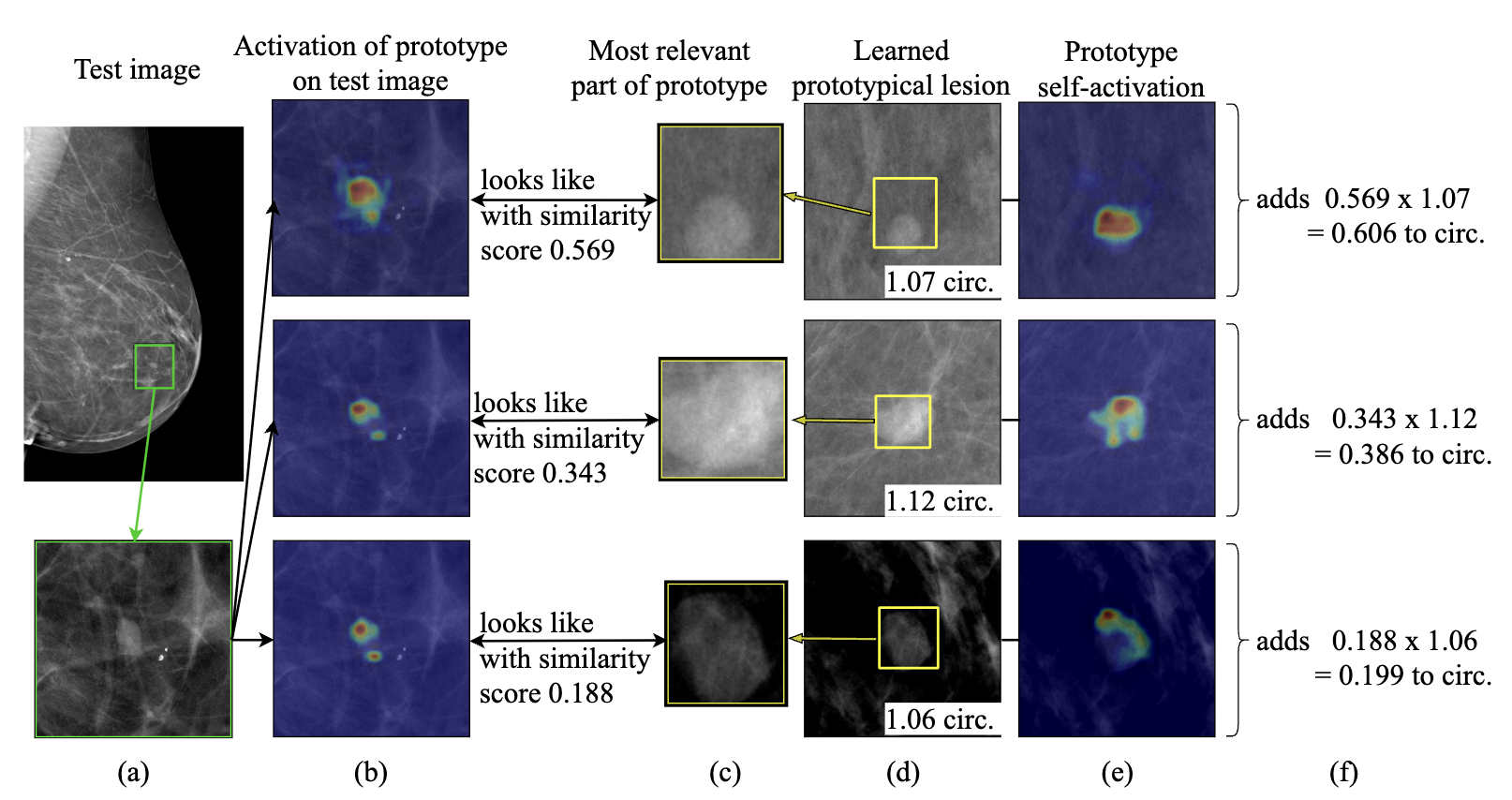}
    \caption{\textbf{Case-based explanation generated by FPN-IAIA-BL}. This circumscribed (circ.) lesion is correctly classified as circumscribed. a, Test images. b, Activation of prototype on test images. c, Most relevant part of prototype. d, Learned prototypical lesion. e, Prototype self-activation. f, Contribution to class score. This visualization format for this figure matches that of \cite{barnett2021case}.}
    \label{fig:preds}
\end{figure*}

The dataset, previously studied in \cite{barnett2021case}, includes 2D digital breast x-rays from patients at the Duke University Health System taken between 2008 and 2018. Data collection was approved by Duke Health IRB and labeled by a fellowship-trained breast imaging radiologist. While IAIA-BL used only the subset of the images that contained a lesion, we also introduce a negative class which consists of images of tissue without lesions. Supplement Section  \ref{app:negdata} details how the data for this class were generated.

The training of FPN-IAIA-BL consists of three stages: (A) a warmup stage, (B) a projection of prototypes, and (C) full network fine-tuning. Because we use a trained VGG-16 backbone from IAIA-BL to construct our FPN, we first freeze the VGG-16 backbone in Stage A to warm up all the other layers. Stage B projects the learned prototype vectors onto a patch from any input image's corresponding feature map in the same fashion as in \cite{barnett2021case, PPNet}. Stage C continues these two stages and unfreezes the VGG-16 backbone to allow for fine-tuning of the full network.


For stages A and C, we minimize the loss function: 
\begin{equation}\ell = \text{CE} + \lambda_1 \ell_{clust} + \lambda_2\ell_{sep}+\lambda_3\ell_{ortho}+\lambda_4\ell_{fine}
\end{equation} 
where cross entropy (CE) penalizes misclassification and $\lambda_1, \lambda_2, \lambda_3, \lambda_4$ are coefficients chosen empirically to balance the cluster ($\ell_{clust}$), separation ($\ell_{sep}$), and orthogonality ($\ell_{ortho}$) losses as defined in \cite{donnelly2022deformable} and fine-annotation loss ($\ell_{fine}$) modified from \cite{barnett2021case}. The modifications to the fine-annotation loss introduce user-configurable coefficients which encourage and penalize the model for activating inside and outside the fine annotations differently for each class pair. Supplement Section \ref{app:fa} details the fine-annotation coefficients. 
 
These loss terms have not previously been combined.

\begin{figure*}
   \centering
   \includegraphics[width=0.90\textwidth]{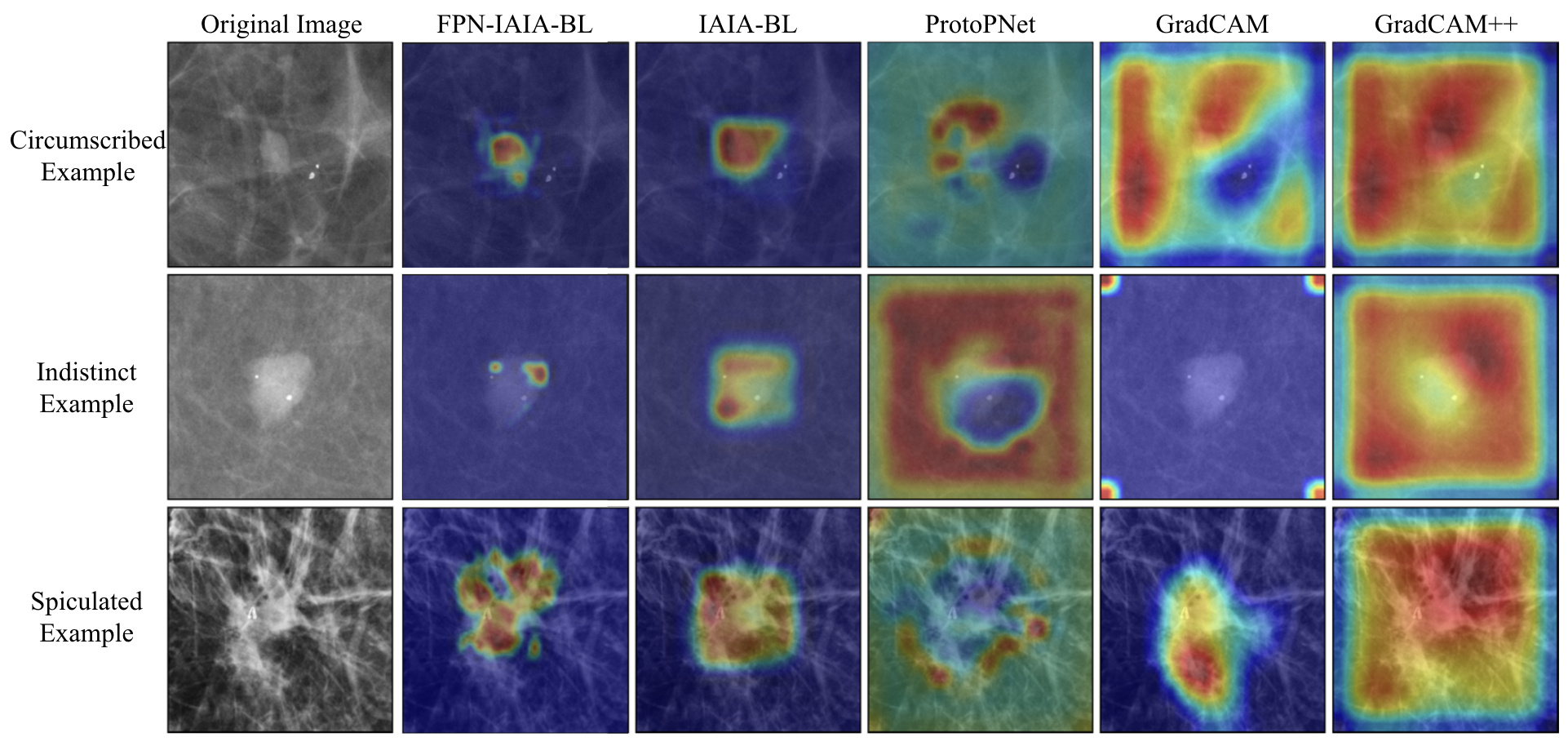}
   \caption{\textbf{FPN-IAIA-BL in comparison to other saliency methods (adapted from \cite{barnett2021case}).} We compare explanations from FPN-IAIA-BL with GradCAM \cite{Selvaraju_2017_ICCV}, GradCAM++ \cite{chattopadhay2018grad}, ProtoPNet \cite{PPNet}, and IAIA-BL \cite{barnett2021case}. GradCam and GradCAM++ are two popular saliency explanation methods, and ProtoPNet and IAIA-BL are case-based explanation methods. The explanations from FPN-IAIA-BL highlight the most important parts of the lesion margin. }
   \label{fig:full}
\end{figure*}

\begin{figure*}[h!]
    \centering
    \begin{subfigure}{0.49\textwidth}
        \centering
        \includegraphics[width=1\textwidth]{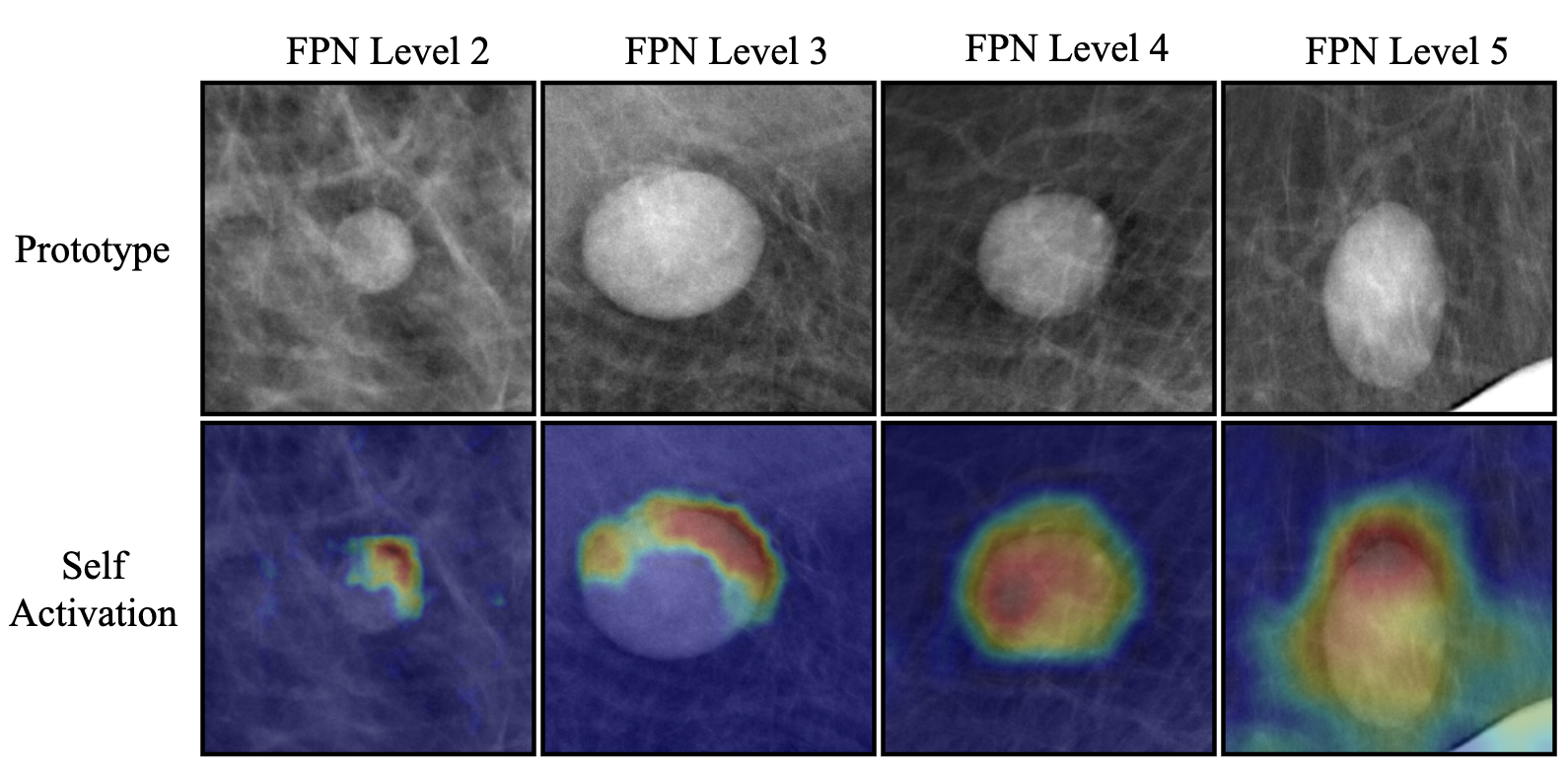}
        \caption{Circumscribed}
        \label{fig:plevel_c}
    \end{subfigure}%
    \vspace{2mm}
    \begin{subfigure}{0.49\textwidth}
        \centering
        \includegraphics[width=1\textwidth]{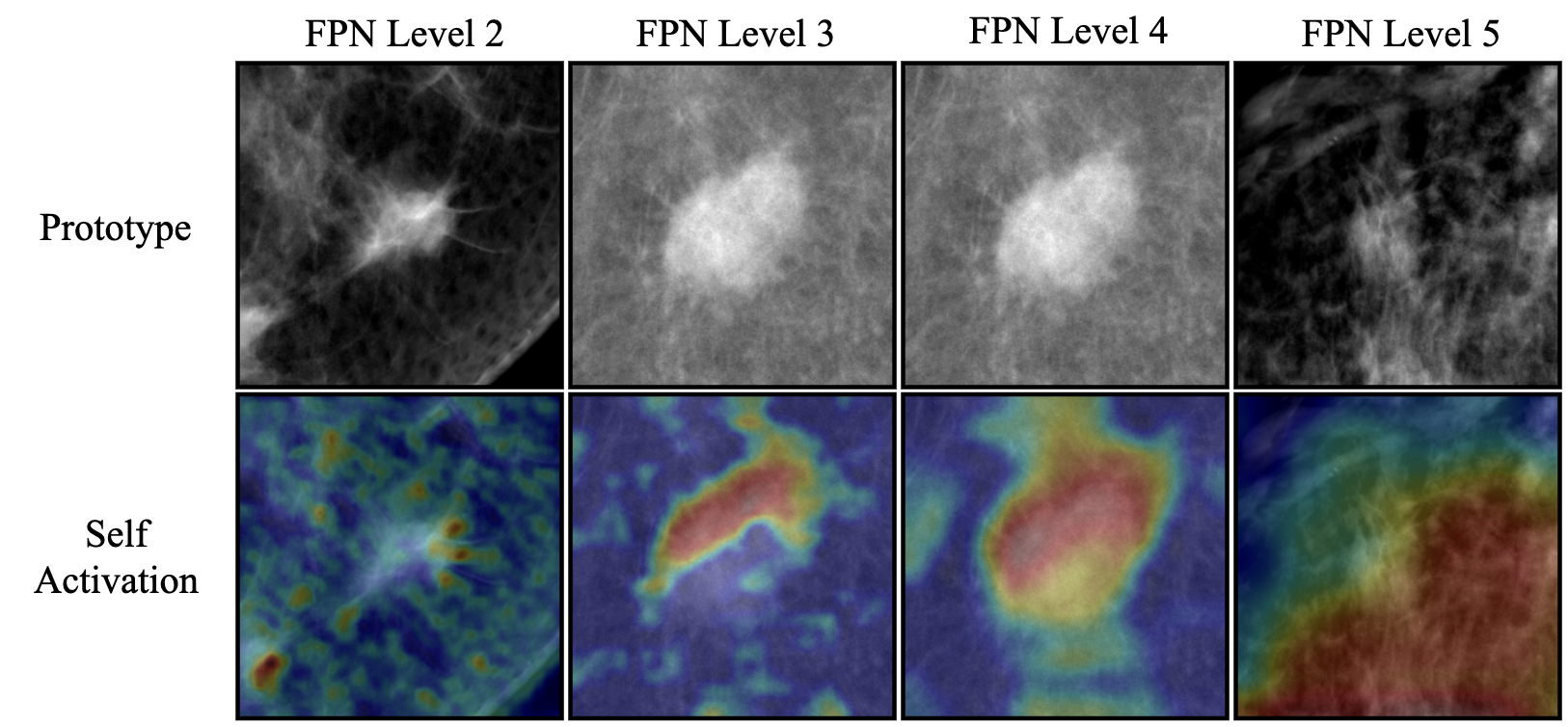}
        \caption{Spiculated}
        \label{fig:plevel_s}
    \end{subfigure}
    \caption{\textbf{Learned prototypes at different FPN-levels.} FPN-level 2 prototypes are more localized because they are learned from the base of the feature pyramid which is a finer-grained feature map while FPN-level 5 prototypes are learned from the top of the feature pyramid, a coarser-grained feature map.} 
    \label{fig:plevels}
\end{figure*}

\section{Experiments and Results}  \label{sec:experiments}
In our experiments, we find that FPN-IAIA-BL is able to learn localized prototypes that achieve acceptable performance.  
An interpretable visual result of the FPN-IAIA-BL is shown in Figure \ref{fig:preds} and is compared to baselines in Figure \ref{fig:full}. The best performing FPN-IAIA-BL model was able to achieve an average AUROC of 0.88 with one-vs-rest AUROC's of 0.865 for circumscribed,  for indistinct, and 0.908 for spiculated margin classes. A further comparison of the performance with IAIA-BL and an uninterpretable baseline (VGG16) is presented in Table \ref{tab:auroc}. The confusion matrix of this model is shown in Supplement Section \ref{app:conf_matrix}. 

\begin{table} [h]
   \centering
   \begin{tabular}{c|c c c c} \hline 
       &  Avg. AUROC & Circ. &  Ind. & Spic\\
       \hline
       FPN-IAIA-BL &  0.88 &  0.87 & 0.86& 0.91\\ 
       IAIA-BL &  0.95 &  0.97 &  0.93 & 0.96\\ 
       VGG16 & 0.95 & 0.95 & 0.94 & 0.95 \\ 
       \hline
   \end{tabular}
   \caption{AUROC metrics for FPN-IAIA-BL as compared to IAIA-BL and the uninterpretable baseline (VGG16).}
   \label{tab:auroc}
\end{table}

As shown in Figure \ref{fig:plevels}, prototypes from each FPN level represent relevant features from multiple scales. FPN-level 2 localizes to the most fine-grained features, and FPN-level 5 activations cover large swaths of the image. 
\textbf{The model successfully learned prototypes at each FPN-level that captured information of different scales. In our application for mass margin classification, FPN-level 3 provided prototypes that activated on the most specific and salient parts of the margin}. 
In other applications, the FPN-level of each prototype can be configured such that the prototypes capture the most relevant scale of information for the application. 
Figure \ref{fig:full} compares the activation maps provided by FPN-IAIA-BL, IAIA-BL, ProtoPNet, GradCAM and GradCAM++. The explanations from FPN-IAIA-BL highlight the most important parts of the lesion margin.

\subsection{Limitations} \label{subsec:next}
While FPN-IAIA-BL consistently produces prototypes for circumscribed and spiculated lesion that our radiology team finds compelling, the prototypes for indistinct margins often activate outside of the lesion. This could be because an indistinct margin is defined as a faded, soft boundary between the lesion and normal tissue, and soft boundaries can occur in healthy breast tissue.
Additionally, 
the AUROC for FPN-IAIA-BL is lower than that of IAIA-BL (0.951 overall) and the uninterpretable baseline (0.947 overall). This is because FPN-IAIA-BL architecture is larger and harder to train than IAIA-BL and the baseline. 

\section{Conclusion}
We presented FPN-IAIA-BL, a novel neural network architecture for multi-scale case-based reasoning. We showed its effectiveness for the task of breast lesion margin classification, creating a model that can articulate more detailed reasoning behind its predictions, improving interpretability.

\section*{Acknowledgements}
This study was supported by National Science Foundation (grant HRD-2222336), Duke TRIPODS CCF-1934964, Duke MEDx: High-Risk High-Impact Challenge, and the Duke Incubation Fund.



\FloatBarrier
\bibliographystyle{plain} 
\bibliography{refs} 

\appendix

\clearpage

\section*{Supplementary Material} 

\section{Confusion Matrix for FPN-IAIA-BL}\label{app:conf_matrix}
\begin{figure} [h!]
    \centering
    \includegraphics[width=0.95\linewidth]{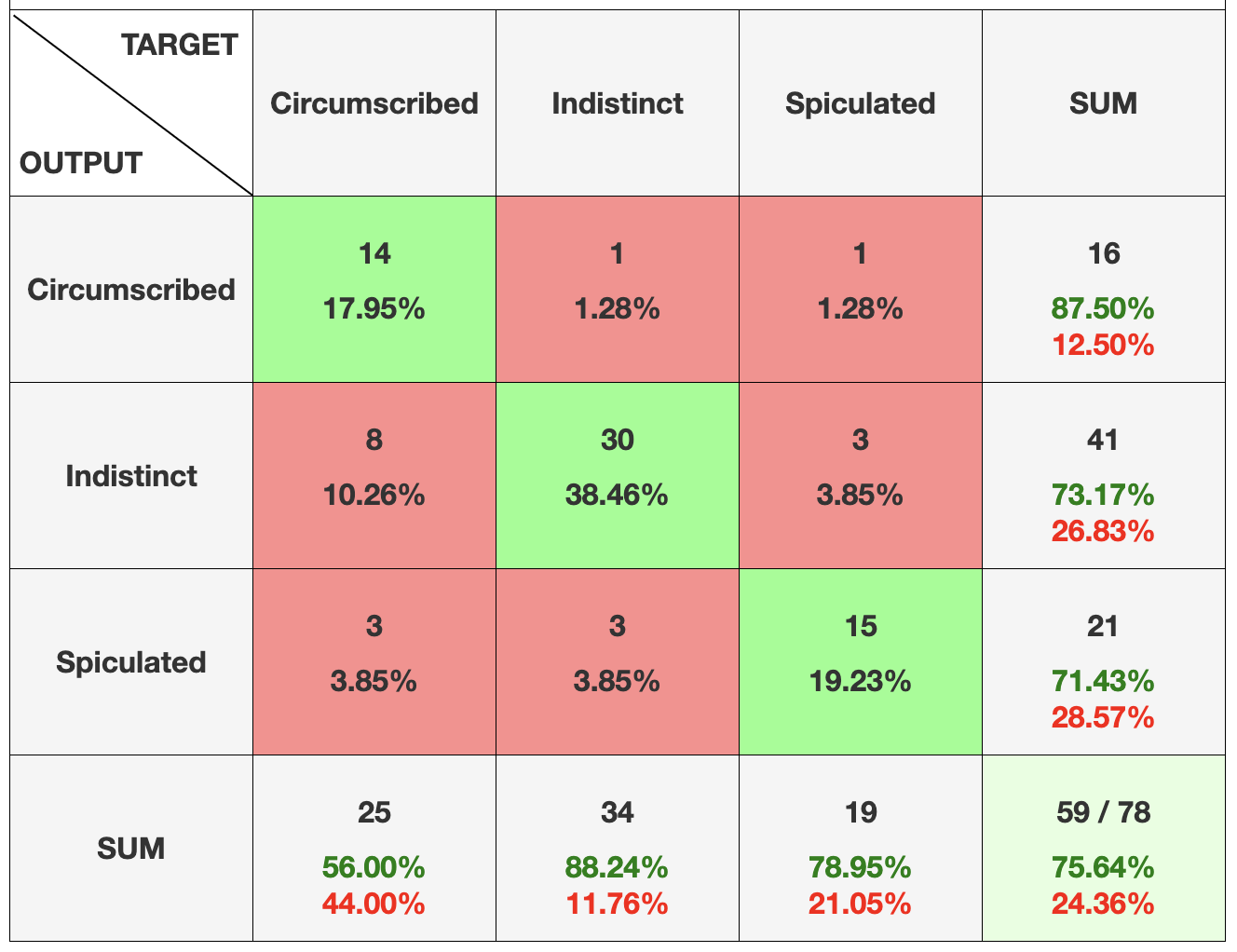}
    \caption{Confusion matrix for predictions on the test dataset.}
    \label{fig:cm}
\end{figure}

Figure \ref{fig:cm} contains the confusion matrix for FPN-IAIA-BL on the test set. It has the highest specificity and lowest sensitivity for the circumscribed class.
\FloatBarrier
\section{Fine Annotation Coefficients}\label{app:fa}
FPN-IAIA-BL introduces fine-annotation coefficients ${\lambda_{\text{in}}^{(y(i), c)}, \lambda_{\text{out}}^{(y(i), c)}}$ which are used in the fine-annotation loss to encourage and penalize the model for activating inside and outside the fine annotations. For example, it is considered ``worse'' for a spiculated prototype to activate on a circumscribed lesion than for a circumscribed prototype to activate on a spiculated lesion. The fine-annotation coefficients designed by board-certified radiologist F.S. are as follow in tables \ref{tab:fa_coeff_in} and \ref{tab:fa_coeff_out}.
\FloatBarrier
\begin{table} [h]
   \centering
   \begin{tabular}{|l|c|c|c|c|l|} \hline 
       \multicolumn{2}{|l|}{ }& \multicolumn{4}{|c|}{Prototype Class}\\ 
       \cline{3-6} 
       \multicolumn{2}{|l|}{ }& Circ. & Ind.  & Spic. &Neg.\\
       \hline
       \multirow{4}{*}{Sample's Class}&Circ.&  1&  1& 1 &1\\ 
       \cline{2-6}
       &Ind.&  1&  1&  1&1\\ 
       \cline{2-6}
       &Spic.&  1&  1&  1&1\\ 
       \cline{2-6}
       &Neg.& 0& 0& 0&0\\ 
       \hline
   \end{tabular}
   \caption{Fine annotation coefficients penalizing the prototypes from class $c_{proto}$ from activating  \textbf{outside} fine annotations for a sample from class $c_{i}$}
   \label{tab:fa_coeff_in}
\end{table}
\begin{table}[h]
   \centering
   \begin{tabular}{|l|c|c|c|c|l|} \hline 
       \multicolumn{2}{|l|}{ }& \multicolumn{4}{|c|}{Prototype Class}\\ 
       \cline{3-6} 
       \multicolumn{2}{|l|}{ }& Circ. & Ind.  & Spic. &Neg.\\
       \hline
       \multirow{4}{*}{Sample's Class}&Circ.&  0&  0& 0&1\\ 
       \cline{2-6}
       &Ind.&  0&  0&  0&1\\ 
       \cline{2-6}
       &Spic.&  1&  1&  0&1\\ 
       \cline{2-6}
       &Neg.& 0& 0& 0&0\\ 
       \hline
   \end{tabular}
   \caption{Fine annotation coefficients penalizing the prototypes from class $c_{proto}$ from activating \textbf{inside} the fine annotations for a sample from class $c_{i}$.}
   \label{tab:fa_coeff_out}
\end{table}
Incorporating the fine-annotation coefficients, the fine-annotation loss is now defined as: 
\begin{equation}
\begin{aligned}
    \ell_{\text{fine}} =  \sum_{i\in D'} \sum_{\mathbf{p}^{(c, l, j)}}  \Big(||\lambda_{\text{in}}^{(y(i),c)} \mathbf{m}_i &\odot \text{PAM}_{i,j} + \\
    &\lambda_{\text{full}}^{(y(i),c)}\text{PAM}_{i,j}||_2\Big)
\end{aligned}
\end{equation} 
where the prototype activation map $\text{PAM}_{i,j}$ is computed by bilinearly upsampling the similarity map $[s_{j,n}]_{n=(1,1)}^{\space\space (\eta_l, \eta_l)}$ for prototype $\mathbf{p}^{(c,l,j)}$ and image $\mathbf{x}_i$ such that it has the same dimensions as the fine-annotation mask $\mathbf{m}_i$.

\section{Negative Class Data} \label{app:negdata}
As discussed in Section \ref{sec:data}, we include a negative class during training to discourage ``classification by elimination.'' The negative class data consist of 5,000 image-mask pairs. The negative class images were created by sampling the full-size mammogram images and cropping to a section without any of the lesion region of interest for each image. We pair each image with a fully negative mask where no region of interest is identified in the mask. For training, we randomly select a subset of 200 negative samples.

\end{document}